\documentclass[a4paper,french]{rnti} 

\usepackage{xcolor}

\usepackage[T1]{fontenc}
\usepackage[utf8]{inputenc}

\usepackage{graphicx, amssymb, amsthm, amsmath}

\titrecourt{Modèles de fondation pour la prévision des séries temporelles}

\nomcourt{M. Laglil et al.}

\titre{Modèles de Fondation et Ajustement : \\
Vers une Nouvelle Génération de Modèles \\
pour la Prévision des Séries Temporelles}

\auteur{Morad Laglil \affil{1},
        Emilie	Devijver \affil{2}\\
        Eric Gaussier  \affil{2}\,
        Bertrand Pracca\affil{3}}

\affiliation{
    \affil{1}\affilsep\affil{2} IMAG
        700 Av. Centrale
        38058 Saint Martin d’Hères, France\\
          morad.laglil@univ-grenoble-alpes.fr,\\
        emilie.devijver@univ-grenoble-alpes.fr\\
        Eric.Gaussier@imag.fr\\

    \affil{3}SAVOYE SASU - Saint-Etienne\\
          bertrand.pracca@savoye.com
 }

\resume{%
Motivés par les récents progrès des grands modèles en traitement automatique du langage naturel, des modèles de fondation ont été développés pour la prévision de séries temporelles en zero-shot, c’est-à-dire sur des jeux de données jamais observés lors du pré-entraînement. Ces modèles, dotés de centaines de millions de paramètres, sont pré-entraînés sur de vastes ensembles de séries temporelles afin d’apprendre des représentations généralisables couvrant la prévision ponctuelle et probabiliste. Ils réduisent ainsi la nécessité de concevoir des modèles spécifiques et d’effectuer un réglage manuel pour chaque tâche.

Dans ce travail, nous présentons un état de l’art des principales architectures et stratégies de pré-entraînement utilisées dans ces modèles, ainsi que leurs méthodes d’optimisation. Nous étudions également l’ajustement de certains modèles de fondation après pré-entraînement, afin d’améliorer leurs performances sur des jeux de données spécifiques. Nos résultats empiriques montrent que cette étape améliore globalement  les capacités de prévision en zero-shot. 
}
\summary{%
Inspired by recent advances in large language models, foundation models have been developed for zero-shot time series forecasting, enabling prediction on datasets unseen during pretraining. These large-scale models, trained on vast collections of time series, learn generalizable representations for both point and probabilistic forecasting, reducing the need for task-specific architectures and manual tuning.

In this work, we review the main architectures, pretraining strategies, and optimization methods used in such models, and study the effect of fine-tuning after pretraining to enhance their performance on specific datasets. Our empirical results show that fine-tuning generally improves zero-shot forecasting capabilities, especially for long-term horizons.
}

\begin{document}

\section{Introduction}

La prévision de séries temporelles est depuis longtemps une tâche fondamentale pour l’industrie, car elle soutient la prise de décision et l’optimisation des processus dans des domaines aussi variés que l’énergie, les transports, la météorologie, l’économie ou encore la distribution. Ce champ constitue également un axe de recherche académique majeur, ayant conduit à de nombreux progrès visant à améliorer la précision des prévisions.

Les approches traditionnelles de prévision reposent principalement sur des modèles paramétriques conçus à partir de l’expertise métier, tels que les modèles autorégressifs (AR, \cite{box2015time}) ou les méthodes de lissage exponentiel \citep{article}. À l’inverse, les approches modernes d’apprentissage automatique — et plus particulièrement l’apprentissage profond — offrent la capacité de modéliser des dynamiques temporelles complexes en apprenant directement à partir des données, sans hypothèses fortes sur leur distribution. Divers modèles, tels que les forêts aléatoires \citep{en15207547} et les réseaux de neurones profonds \citep{SALINAS20201181}, ont ainsi été développés pour améliorer la précision des prévisions et mieux capturer la diversité des comportements temporels observés dans les séries réelles. 

Très récemment, les grands modèles de fondation, pré-entraînés sur d’importants volumes de données, se sont imposés comme une avancée majeure dans la prévision de séries temporelles. Bénéficiant de la disponibilité croissante des données et de la puissance de calcul, ces modèles exploitent des biais inductifs reflétant les propriétés structurelles des séries temporelles afin d’apprendre des représentations généralisables. Cette capacité leur permet de s’adapter à une grande variété de contextes, y compris à des situations inédites (zero-shot).
Inspirés des succès observés dans le traitement automatique du langage, ces modèles peuvent également être ajustés (fine-tuned) sur des jeux de données spécifiques, permettant ainsi de transférer et de spécialiser les connaissances acquises lors du pré-entraînement. Toutefois, cette question d’ajustement demeure très peu explorée dans la littérature sur les modèles de fondation pour la prévision.
Dans cet article, nous nous concentrons sur deux axes principaux :
\begin{itemize}
\item un état de l’art des modèles de fondation pour la prévision de séries temporelles.
\item une étude expérimentale de l’ajustement de certains  modèles sur un large éventail de jeux de données couvrant différents domaines, fréquences et horizons de prévision. A notre connaissance, il s'agit de la première étude d'ajustement de cette envergure.
\end{itemize}

Le plan de ce papier est le suivant. Dans la Section \ref{sec:framework}, nous formalisons le problème de prévision des séries temporelles. La Section \ref{sec:bib} présente un état de l'art détaillé, la Section \ref{sec:expes} expose les résultats des expériences sur l'ajustement, et la Section \ref{sec:conc} discute  certains défis avant de conclure.

\section{Problème de prévision des séries temporelles} \label{sec:framework}
La prévision des séries temporelles consiste à prédire les valeurs futures d'une série indexée par le temps à partir des observations passées et d'autres variables exogènes. Formellement, soit 
$y_t \in \mathbb{R}^D$ le vecteur d’observations à l’instant $t$, et considérons une série temporelle, notée  $\{y_t\}_{t=1}^T$.  
La tâche de prévision consiste à estimer les valeurs futures $\{y_{T+1}, \ldots, y_{T+H}\}$ pour un horizon $H > 0$, à partir des observations passées (contexte) $\{y_1, \ldots, y_T\}$ et, éventuellement, de $n_{\text{cov}}$ covariables externes $\{z_1,\ldots, z_T\}$.  

Nous définissons la fonction de prévision comme une application :
\begin{align*}
f : \mathbb{R}^{T \times D} \times \mathbb{R}^{T \times n_{\text{cov}}} &\longrightarrow \mathbb{R}^{H \times D}\\
(\{y_t\}_{t=1}^T, \{z_t\}_{t=1}^T) &\mapsto \{y_{T+1}, \ldots, y_{T+H}\}.
\end{align*}
Dans le cas d’une prévision probabiliste, la tâche consiste simplement à prédire des quantiles de la distribution future au lieu d’une valeur unique.


\section{Modèles de fondation pour la prévision des séries temporelles}\label{sec:bib}

Dans cette section, nous présentons l’état de l’art relatif à ces modèles, en mettant l’accent sur trois composantes clés :
l’\textbf{architecture} (Section~\ref{sec:archi}), qui définit la structure du réseau et la technologie sous-jacente ;
le \textbf{pré-entraînement} (Section~\ref{sec:preentr}), qui consiste à exposer le modèle à de vastes ensembles de données via un apprentissage auto-supervisé ;
et l’\textbf{adaptation} (Section~\ref{sec:ajus}), qui regroupe les différentes stratégies d’ajustement.
Ces trois dimensions constituent un cadre d’analyse unifié, permettant à la fois de comparer, classer et mettre en œuvre les modèles de fondation de prévision. Pour décrire ces modèles, nous nous appuierons sur la Figure \ref{encodeur}, qui illustre les blocs clés des architectures neuronales classiques. Nous détaillons plusieurs modèles, présentés de façon chronologique, ayant obtenu les meilleures performances sur le benchmark \textit{GIFT-Eval} \citep{aksu2024gifteval} et illustrant une diversité d'approches selon les trois composantes précédemment décrites.

\subsection{Architecture}\label{sec:archi}
\subsubsection{Modèles basés sur les Transformers}

\begin{figure}[t] 
\includegraphics[width=13cm]{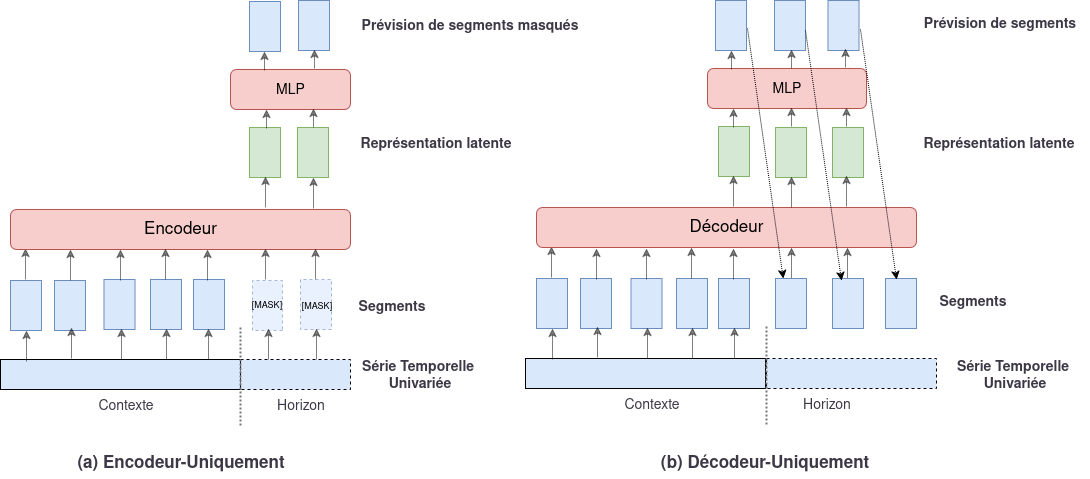} \caption{Illustration des styles d’architectures : (a) encodeur uniquement avec masquage et (b) décodeur uniquement, avec structure autorégressive. En bleu : la série temporelle d’entrée et ses segments, en rouge : les couches neuronales, et en vert : les représentations latentes des segments. MLP désigne les couches neuronales linéaires utilisées pour générer la prévision à partir des représentations latentes.} \label{encodeur} 
\end{figure}

La majorité des modèles de fondation pour la prévision des séries temporelles repose sur l'architecture \textbf{Transformer} \citep{NIPS2017_3f5ee243}, un type de modèle initialement conçu pour le traitement du langage naturel. Leur principale innovation réside dans le \textbf{mécanisme d’attention}, qui permet au modèle d’accorder une importance variable aux différentes parties de la séquence d’entrée. Ce mécanisme offre la capacité d’apprendre des dépendances à longue portée entre les éléments de la série, ce que les architectures précédentes, comme les réseaux récurrents, peinaient à faire.

\cite{ansari2024chronos} ont développé un modèle appelé \textbf{Chronos}, basé sur une architecture Transformer de type \textit{encodeur-décodeur} (combinant les deux architectures de la Figure \ref{encodeur}). L’encodeur analyse la série temporelle passée pour en extraire des représentations riches, tandis que le décodeur génère les prévisions de manière \textit{autoregressive} à partir de ces représentations. Ce modèle reprend les principes du traitement de texte, mais introduit une idée originale : la représentation des séries temporelles sous forme de jetons.  
Concrètement, chaque valeur de la série est convertie en un jeton correspondant à l’intervalle prédéfini auquel elle appartient. À la sortie, les jetons prédits par le modèle sont reconvertis en valeurs réelles, par exemple en leur associant la borne inférieure de l’intervalle correspondant. Une limite de cette approche est que la plage de prédiction reste fixe, car les intervalles sont prédéfinis, ce qui la rend moins adaptée aux séries temporelles présentant des tendances fortes. 

Une autre approche pour convertir une série temporelle en unités d’information (ou jetons) exploitables par les Transformers consiste à la segmenter en plusieurs sous-séquences de taille fixe, appelées segments. En divisant les longues séries temporelles en segments plus courts, le modèle devient capable de mieux capturer les motifs locaux au sein de la séquence, améliorant ainsi sa compréhension des dynamiques complexes qu’elle renferme. De plus, cette segmentation contribue à réduire la complexité du modèle Transformer en diminuant le nombre total de jetons à traiter.
\cite{10.5555/3692070.3692474} ont appliqué cette idée dans le modèle \textbf{TimesFM}, basé sur une architecture Transformer de type \textit{décodeur-uniquement} (Figure \ref{encodeur}(b); dans ce cas, le décodeur est un Transformer). Ce dernier repose sur un mécanisme d’attention causale, où chaque segment ne peut accéder qu’aux segments précédents, sans jamais utiliser d’informations futures. Ainsi, le modèle apprend à effectuer des prévisions en extrayant uniquement les dépendances pertinentes issues du passé.
Le modèle adopte un schéma de génération \textit{autoregressive} lors de l’inférence, produisant successivement les segments de sortie. La taille des segments prédits est supérieure à celle des segments d’entrée, ce qui permet d’assurer une prévision à long terme tout en réduisant le nombre d’itérations nécessaires. Cette conception s’appuie sur des travaux antérieurs \citep{Zeng2022AreTE}, qui ont montré que la prévision de l’ensemble de l’horizon en une seule étape à partir du passé offre généralement une meilleure précision que les approches autoregressives itératives sur ces modèles.

\textbf{Moirai} \citep{woo2024moirai} reprend le principe de segmentation des séries temporelles, mais introduit cette fois une taille de segment variable: plus grande pour les séries à haute fréquence (par exemple à la seconde ou à la minute) afin de réduire le coût quadratique de l’attention, plus petite pour les séries à basse fréquence (jour, mois) pour mieux exploiter la capacité de représentation des Transformers.
Pour cela, Moirai apprend plusieurs couches de projection d’entrée et de sortie associées à différentes tailles de segments, la sélection manuelle de la taille appropriée dépendant des caractéristiques fréquentielles de la série.
\\
Le cœur du modèle repose sur un Transformer à encodeur uniquement (Figure \ref{encodeur}(a), dans ce cas l'encodeur est un Transformer), utilisant une technique de masquage : un jeton fictif \texttt{[mask]} remplace les segments situés dans l’horizon de prévision. Ce jeton possède sa propre représentation apprenable, qui est ensuite décodée via une projection de sortie pour générer la prévision correspondante, ce qui permet d’éviter l’autoregressivité lors de l’inférence.\\
Un autre défi pour les modèles de fondation en prévision réside dans leur difficulté à traiter des séries temporelles multivariées de dimensions arbitraires. La plupart des Transformers existants supposent une indépendance entre les variables, car le nombre de dimensions peut varier et n’est pas connu à l’avance.
Pour surmonter cette limitation, Moirai adopte une approche consistant à aplatir la série multivariée afin de la traiter comme une séquence unique. Cette stratégie s’accompagne de l’introduction d’encodages des indices de variables, permettant au modèle de distinguer les différentes dimensions lors du calcul de l’attention. 

\textbf{TOTO} \citep{cohen2025this}, récemment proposé, est un décodeur (Figure \ref{encodeur}(b)) basé sur un Transformer qui adopte une approche alternative pour le traitement des séries temporelles multivariées. Chaque série est d’abord découpée en plusieurs segments, puis un mécanisme d’attention à deux niveaux est appliqué : le premier capture les dépendances temporelles au sein de chaque série univariée, tandis que le second modélise les interactions entre séries via une attention dimensionnelle \citep{zhang2023crossformer}. Cette architecture permet à TOTO de représenter efficacement à la fois les relations temporelles et inter-variables.

\subsubsection{Modèles basés sur les réseaux récurrents}
Les réseaux de neurones récurrents, en particulier les LSTM \citep{10.1162/neco.1997.9.8.1735}, ont été largement utilisés pour la prévision de séries temporelles \citep{SALINAS20201181}. Le premier, et à ce jour le seul, modèle de fondation reposant sur une architecture récurrente est \textbf{TiRex} \citep{auerTiRexZeroShotForecasting2025}. Il s’agit d’un décodeur (Figure \ref{encodeur}(b)) basé sur le sLSTM \citep{beck:24xlstm}, une extension du LSTM classique qui introduit une activation exponentielle à la place de la fonction sigmoïde, laquelle contraint la sortie à l’intervalle [0, 1]. Cette modification confère au réseau une expressivité accrue.
TiRex segmente les séries temporelles en segments. Contrairement aux décodeurs autoregressifs, qui génèrent les prévisions segment par segment (par exemple lors de l’inférence) en réinjectant chaque sortie prédite à l’entrée — ce qui accentue l’accumulation d’erreurs —, TiRex considère les segments futurs comme des valeurs manquantes, évitant ainsi la propagation d’erreurs et améliorant la stabilité des prévisions à long terme.

\subsection{Pré-entraînement}\label{sec:preentr}
Le pré-entraînement constitue une étape initiale et essentielle dans la construction des modèles de fondation pour les séries temporelles. Les connaissances acquises au cours de cette phase permettent aux modèles de mieux généraliser à travers différents contextes.
Cependant, la diversité des styles architecturaux et des types de prévision visés (par exemple, encodeurs avec masquage, décodeurs autoregressifs, ou encore prévision probabiliste versus ponctuelle) conduit à une grande variété de mécanismes de pré-entraînement et de stratégies d’optimisation lors de la conception et du développement de ces modèles.

 Le pré-entraînement est effectué de manière auto-supervisée, sans nécessiter d’annotation des données. Chronos \citep{ansari2024chronos} est pré-entraîné en minimisant la fonction d’entropie croisée sur les jetons générés de manière autoregressive pour un horizon de prédiction fixe. Le modèle apprend à prédire une distribution sur l’ensemble des jetons définis, ce qui lui permet de générer des prévisions probabilistes (par exemple pour les quantiles 0.1, 0.2, …, 0.9).

TimesFM \citep{10.5555/3692070.3692474} et TiRex \citep{auerTiRexZeroShotForecasting2025} partagent la même idée de pré-entraînement. 
Pour une séquence de segments d’entrée, le modèle est optimisé pour prédire le segment suivant à partir des précédents 
en minimisant la fonction de perte de quantile définie par:
\begin{equation}
 \mathcal{L}(y_{T+1:T+H}, \hat{y}^q_{T+1:T+H}; Q) = \frac{1}{|Q|\, H} \sum_{t=T+1}^{T + m_{\text{out}}} \sum_{q \in Q} \big(q - \mathbf{1}_{\{y_t - \hat{y}^q_t < 0\}}\big)\, (y_t - \hat{y}^q_t),
\end{equation}

où $Q$ désigne l’ensemble des niveaux de quantile, $\hat{y}^q_t$ est la prédiction du modèle pour le niveau de quantile $q$ à l'instant $t$ et $y_t$ est la valeur réelle. \\
TiRex utilise, lors de l’inférence, la prédiction multi-segment en traitant les segments futurs comme des données manquantes (en limitant l’accès à ces parties lors du calcul de l’attention) afin d’éviter l’autoregressivité. Pour simuler ce comportement lors du pré-entraînement, le modèle applique un masquage de segments contigus en masquant aléatoirement un ensemble de segments consécutifs dans la série d’entrée, reproduisant ainsi un scénario d’inférence.

Le modèle Moirai \citep{woo2024moirai} est pré-entraîné en maximisant la log-vraisemblance d’une distribution de mélange paramétrique 
combinant plusieurs lois (Student-t, binomiale négative, log-normale et normale) afin de modéliser différents types de données temporelles. 
L’entraînement est réalisé sur un large corpus appelé  \textit{LOTSA} selon une distribution hiérarchique de données et de tâches. Cette stratégie permet au modèle d’apprendre à partir de fenêtres de contexte et d’horizon de longueurs variables, ainsi que de séries multi-domaines et multivariées,
ce qui renforce sa capacité de généralisation.

\subsection{Apprentissage par transfert: Ajustement sur des tâches spécifiques}\label{sec:ajus}

L’ajustement, ou \textit{fine-tuning}, des modèles de fondation est une technique d’apprentissage par transfert qui consiste à exploiter les connaissances acquises lors du pré-entraînement sur de vastes ensembles de données pour améliorer les performances sur une tâche cible.  
Cette approche est particulièrement cruciale dans le domaine de la prévision temporelle, où les jeux de données disponibles sont souvent de petite taille (par exemple, une seule série temporelle), limitant ainsi l’efficacité des modèles d’apprentissage automatique classiques qui nécessitent un grand nombre d’exemples pour bien fonctionner.  
En réutilisant les représentations apprises sur d’autres domaines, l'ajustement permet de contourner cette contrainte. \cite{Yuqietal-2023-PatchTST} a notamment démontré les effets positifs de l’ajustement de modèles pré-entraînés basés sur des architectures de type Transformer. Deux approches principales peuvent être envisagées :  
(i) ajuster l’ensemble des paramètres du modèle en poursuivant l’entraînement sur la tâche cible ; ou  
(ii) n’ajuster que certaines parties spécifiques du modèle — comme les couches de projection linéaires d’entrée et de sortie, ou certaines couches du Transformer — tout en conservant les autres paramètres figés.


\section{Expérimentations}\label{sec:expes}

\subsection{Données utilisées et modèles expérimentés}
Nous menons notre étude d’ajustement sur l’ensemble de jeux de données constituant la partie train/test du benchmark GIFT-Eval \citep{aksu2024gifteval}, un vaste cadre d’évaluation dédié aux modèles de prévision. Cette section du benchmark comprend 15 jeux de données univariées et 8 multivariées, couvrant 7 domaines et 10 fréquences différentes, pour un total de \mbox{144\,000} séries temporelles et 177 millions d’observations.
Au total, 92 configurations uniques, combinant jeu de données, fréquence et horizon de prévision (court, moyen et long terme), sont considérées, offrant ainsi une évaluation exhaustive et diversifiée des capacités d’ajustement des modèles de fondation de prévision. Une description complète du jeu de données est fournie dans \citet{aksu2024gifteval} (Voir l’Annexe pour plus de détails sur les jeux de données utilisés).

Le choix des modèles repose sur deux critères : leurs performances (MWQL) sur le benchmark GiftEval et la disponibilité du code de reproduction. Parmi ceux dont le code d’entraînement est accessible, nous avons ainsi retenu le meilleur modèle de la section zero-shot et les deux meilleurs modèles de la section pre-trained du tableau de bord GiftEval \footnote{https://huggingface.co/spaces/Salesforce/GIFT-Eval} à la date de notre étude. Cette sélection couvre les principales familles d’architectures de modèles de fondation pour la prévision : Moirai-1.1-R-Large (310M paramètres, Encoder-only), Chronos-Bolt-Base (205M paramètres, Encoder–Decoder) et TimesFM-2.0 (500M paramètres, Decoder-only). 

Dans notre étude, nous comparons les performances de ces modèles ajustés (Full Fine-Tuning, FFT) à celles de leurs versions zero-shot. Il est toutefois important de noter que Chronos-Bolt-Base et TimesFM-2.0 ont été pré-entraînés sur une partie limitée des ensembles de train/test du benchmark GIFT-Eval. Par conséquent, pour certains jeux de données de notre benchmark, l’évaluation de ces modèles en mode zero-shot n’est pas entièrement fiable car ils ont potentiellement été exposés à une partie des données de test durant le pré-entraînement. Cependant, dans notre étude, l’impact de cette contamination Pré-entraînement/Test demeure limité puisque nous comparons chaque modèle zero-shot uniquement à sa propre version ajustée, les deux partageant le même jeu de données de pré-entraînement. Cet effet devient en revanche plus significatif lorsqu’on compare différents modèles en zero-shot, comme le souligne \cite{aksu2024gifteval}.
\subsection{Protocole expérimental}

Nous avons choisi d’ajuster entièrement les modèles (ajuster l’ensemble de leurs paramètres) sur chaque tâche cible, car leur taille relativement modeste — au maximum 500 millions de paramètres — offre une plus grande flexibilité pour l’ajustement.
Étant donné le grand nombre d’expériences à effectuer, nous n’effectuons pas la recherche des hyperparamètres optimaux pour chaque tâche ; nous utilisons plutôt les mêmes hyperparamètres pour un modèle donné sur l’ensemble des tâches.
Le Tableau \ref{tab:params} présente les hyperparamètres employés pour l’ajustement de chaque modèle, pour lesquels nous retenons les valeurs par défaut. L’étude de la sélection d’hyperparamètres optimale est laissée à de futurs travaux.
\paragraph{Métriques d’évaluation.}
Nous évaluons la performance des modèles de prévision à l’aide de deux métriques complémentaires : l’erreur moyenne absolue en pourcentage (\textit{Mean Absolute Percentage Error}, MAPE) pour la prévision ponctuelle, et le score de probabilité classé continu (\textit{Continuous Ranked Probability Score}, CRPS) pour la prévision probabiliste.

La MAPE mesure la différence moyenne absolue en pourcentage entre les valeurs prédites $\hat{y}_t$ (nous utilisons la prédiction de la médiane) et les valeurs réelles $y_t$, et est défini par:
\begin{equation}
\text{MAPE} = \frac{100}{H} \sum_{t=T+1}^{T+H} \left| \frac{y_t - \hat{y}_t}{y_t} \right|,
\end{equation}
où $H$ représente la taille de l'horizon.

Le CRPS est une métrique utilisée en prévision probabiliste pour évaluer la précision des fonctions de répartition $F$ prédites par rapport aux valeurs observées. 
Pour une valeur réelle $y$, le CRPS est défini comme :
\begin{equation}
\text{CRPS}(F, y) = \int_0^1 2 \, \Lambda_\alpha(F^{-1}(\alpha), y) \, d\alpha,
\end{equation}
où la perte quantile $\Lambda_\alpha(q, y)$ est donnée par : $\Lambda_\alpha(q, y) = (\alpha - \mathbf{1}_{\{y < q\}})(y - q)$,
avec $\mathbf{1}_{\{y < q\}}$ l’indicateur valant 1 si $y < q$, et 0 sinon.

\medskip

En pratique, le calcul intégral du CRPS peut être coûteux en termes de complexité computationnelle, et nous l'approximons  par une somme discrète sur un ensemble fini de niveaux de quantiles. 
Cette approximation, souvent appelée perte quantile pondérée moyenne (\textit{Mean Weighted Quantile Loss}, \cite{aksu2024gifteval}), est définie par :

\[
{\mathrm{MWQL}} = \frac{2}{K} \sum_{k=1}^{K}  \frac{\sum_{t=T+1}^{T+H} \Lambda_{\alpha_k}(\hat{q}_t(\alpha_k), y_t)}{\sum_{t=T+1}^{T+H} |y_t|} ,
\]

où $K$ est le nombre de niveaux de quantiles, 
$Q = \{\alpha_1, \alpha_2, \ldots, \alpha_K\}$ l'ensemble des quantiles retenus 
avec $\alpha_k = 0.1k$ pour $k \in \{1, 2, \ldots, 9\}$ lorsque $K = 9$, 
$\hat{q}_t(\alpha)$ le quantile prédit au niveau $\alpha$ à l'instant $t$, 
$y_t$ la valeur observée réelle, 
et $\Lambda_\alpha(\hat{q}_t(\alpha), y_t)$ la perte quantile, 
laquelle caractérise la discordance entre le quantile prédit et l'observation 
via une pénalisation asymétrique dépendant de $\alpha$.

\begin{table}[t]
\begin{center}
\footnotesize
\setlength{\tabcolsep}{4pt} 
\begin{tabular}{lccc}
\hline
\textbf{Paramètre} & \textbf{Moirai-1.1-R-Large} & \textbf{Chronos-Bolt-Base} & \textbf{TimesFM-2.0}  \\
\hline
Taux d'apprentissage  & $1\times10^{-7}$ & $1\times10^{-5}$& $1\times10^{-6}$\\
Batch size  & 64 & 256 & 128 \\
Nombre max. d'itérations de gradient & 4000 & 4000 & 4000 \\
Fonction optimisée & Vraisemblance de loi de mélange & Perte Quantile & Perte Quantile \\
\hline
\end{tabular}
\caption{Hyperparamètres et fonction de perte utilisée pour l’ajustement des modèles de fondation.} 
\label{tab:params}
\end{center}
\end{table}

Il convient de noter que la fonction de perte de quantile et la vraisemblance du modèle de mélange (utilisée dans l'apprentissage des modèles utilisés, voir Tableau \ref{tab:params}) sont directement liées au $\mathrm{MWQL}$. L’optimisation de ces fonctions se traduit donc par une réduction du $\mathrm{MWQL}$, améliorant la qualité des prévisions probabilistes.

Comme la magnitude des métriques d’évaluation peut varier d’un jeu de données à l’autre,nous normalisons les métriques de chaque modèle en les divisant par celles d’un modèle de référence (ici, Seasonal Naive) : $\frac{\mathrm{Metric}_{\text{model}}}{\mathrm{Metric}_{\text{ref}}}$. 
Les scores relatifs obtenus sont ensuite agrégés à l’aide de la moyenne arithmétique. Nous notons $\mathrm{MWQL}_n$ et $\mathrm{MAPE}_n$ ces métriques relatives.

\subsection{Résultats et Analyse}
Nous présentons les résultats selon deux axes d’analyse.
Premièrement, nous agrégeons les performances en fonction de la taille des jeux de données — petits, moyens et grands (définis à partir des quantiles 0.33 et 0.66 de la distribution du nombre d’observations par jeu de données) — afin d’évaluer l’impact de la taille de données sur la qualité de l’ajustement.
Deuxièmement, pour obtenir une vision plus détaillée des performances, nous agrégeons les résultats selon le type d’horizon de prévision (court, moyen et long terme, tel que défini dans \cite{aksu2024gifteval}) au sein de chaque domaine d’application (énergie, ventes, finance, etc.).

\begin{table}[t]
\centering
\small
\resizebox{\textwidth}{!}{\begin{tabular}{ll |cc|cc|cc}
\hline
\textbf{Modèle} & \textbf{Config} 
& \multicolumn{2}{c|}{\textbf{Petit}} 
& \multicolumn{2}{c|}{\textbf{Moyen}} 
& \multicolumn{2}{c}{\textbf{Grand}} \\
\cline{3-8}
& & $\mathrm{MWQL}_n$ & $\mathrm{MAPE}_n$ & $\mathrm{MWQL}_n$ & $\mathrm{MAPE}_n$ & $\mathrm{MWQL}_n$ & $\mathrm{MAPE}_n$ \\
\hline
Moirai-1.1-R-L & Zero-shot        & 0.634 & 0.745 & 0.541 & 0.862 & 0.353 & \textbf{0.939} \\
Moirai-1.1-R-L & FFT & \textbf{0.544±0.007} & \textbf{0.627±0.009} & \textbf{0.414±0.000} & \textbf{0.763±0.002} & \textbf{0.310±0.000} & 1.056±0.013 \\
\hline
Chronos-Bolt-Base & Zero-shot & 0.594 & 0.768 & 0.513 & 1.111 & 0.311 & \textbf{0.895} \\
Chronos-Bolt-Base & FFT & \textbf{0.559±0.020} & \textbf{0.683±0.018} & \textbf{0.431±0.001} & \textbf{0.687±0.012} & \textbf{0.309±0.005} & 0.991±0.034\\
\hline
TimesFM-2.0 & Zero-shot  & 0.593 & 0.736 & 0.476 & \textbf{0.848} & 0.344 & \textbf{1.178} \\
TimesFM-2.0 & FFT & \textbf{0.591±0.002} & \textbf{0.735±0.005} & \textbf{0.468±0.003} & 0.893±0.011 & \textbf{0.328±0.002} & 1.233±0.033 \\
\hline
\end{tabular}
}
\caption{Moyennes de $\mathrm{MWQL}_n$ et $\mathrm{MAPE}_n$ selon la taille des jeux de données (petit, moyen, grand), avec distinction entre les modes Zero-shot et en full fine-tuning (FFT).}
\label{tab:sizeagr}
\end{table}

\paragraph{Taille des jeux de données (Voir Tableau \ref{tab:sizeagr}):}Concernant le $\mathrm{MWQL}_n$, qui est le critère optimisé pendant l'ajustement, nous observons une amélioration systématique après ajustement pour l’ensemble des modèles et des tailles de jeux de données.
Pour Moirai et Chronos, cette amélioration est particulièrement marquée sur les jeux de données de petite et moyenne taille. Cela s’explique par le fait qu’avec un nombre d’itérations de gradient fixe pour tous les ensembles, ces modèles ont pu parcourir plusieurs fois les jeux de données de taille réduite. Ainsi, ils ont davantage consolidé l’apprentissage des motifs récurrents présents dans ces ensembles. En revanche, pour les grands jeux de données, les modèles n’ont pas eu l’occasion de voir la totalité des exemples, ce qui limite l’effet de l’ajustement.

À l’inverse, TimesFM présente une amélioration du $\mathrm{MWQL}_n$ plus significative sur les grands jeux de données. Cela peut s’expliquer par sa capacité à mieux exploiter la diversité des exemples, plutôt que la fréquence d’exposition aux mêmes données.

Enfin, en ce qui concerne la $\mathrm{MAPE}_n$, celle-ci s’améliore globalement après ajustement, à l’exception des grands jeux de données où une dégradation est observée. Cette baisse de performance peut s’expliquer par une plus grande variabilité des valeurs cibles et par le fait que la $\mathrm{MAPE}_n$ n’a pas été  optimisée lors de l’ajustement.

\paragraph{Domaines et horizons (voir Tableaux \ref{tab:part2} et \ref{tab:part1}) :}
Nous constatons que les performances obtenues après ajustement varient selon le modèle et le domaine étudié.
Concernant le $\mathrm{MWQL}_n$, le modèle Moirai ajusté parvient à améliorer les résultats dans la grande majorité des cas. L’amélioration est particulièrement marquée pour les horizons moyens et longs dans les domaines du Transport, du Web/CloudOps et de l’Énergie, où la version en zero-shot montre des performances limitées. Cela suggère que l’ajustement constitue une étape essentielle pour renforcer la qualité des prévisions à moyen et long terme.

Le modèle Chronos-Bolt-Base ajusté surpasse généralement sa version zero-shot en terme de $\mathrm{MWQL}_n$ dans la plupart des domaines, à l’exception du domaine Énergie, où une légère dégradation est observée. Cette baisse peut s’expliquer par un surapprentissage dû à la nature spécifique de ce domaine, souvent marqué par un changement de distribution entre les ensembles d’entraînement et de test \citep{wang2023timemixer}. Dans ce cas, un ajustement plus fin serait nécessaire — notamment en optimisant les hyperparamètres (tels que le taux d’apprentissage et le nombre maximal d’itérations de gradient) et en appliquant une régularisation adaptée pour limiter le surapprentissage et améliorer la généralisation.
Quant à TimesFM-2.0 ajusté, l’amélioration du $\mathrm{MWQL}_n$ est quasi systématique, à l’exception du domaine Web/CloudOps, où la version zero-shot reste supérieure.

En revanche, pour la $\mathrm{MAPE}_n$, les performances des modèles ajustés se dégradent dans plusieurs domaines et horizons, contrairement à l’amélioration observée pour le $\mathrm{MWQL}_n$, notamment pour Chronos-Bolt-Base et TimesFM. Cette dégradation peut s’expliquer par le fait que la fonction de perte de quantile utilisée lors de l’ajustement privilégie l’optimisation du $\mathrm{MWQL}_n$, parfois au détriment de la $\mathrm{MAPE}_n$, qui n’est représentée que par un seul quantile (la médiane) dans cette fonction.

\begin{table}[t]
\centering
\small
\resizebox{\textwidth}{!}{
\begin{tabular}{ll|ccc|ccc|ccc}
\hline
\textbf{Modèle} & \textbf{Config} & \multicolumn{3}{c|}{\textbf{Transport}} & \multicolumn{3}{c|}{\textbf{Nature}} & \multicolumn{3}{c}{\textbf{Web/CloudOps}} \\
\cline{3-11}
& & Short & Medium & Long & Short & Medium & Long & Short & Medium & Long \\
\hline
&& \multicolumn{9}{c}{\textbf{$\mathrm{MWQL}_n$}} \\
\hline
Moirai-1.1-R-L & Zero-shot 
& 0.347 & 0.147 & 0.132 & 0.431 & 0.229 & 0.186 & 0.699 & 0.653 & 0.686 \\

Moirai-1.1-R-L & FFT 
& \textbf{0.340 ± 0.000} & \textbf{0.130 ± 0.000} & \textbf{0.114 ± 0.000} & \textbf{0.398 ± 0.002} & \textbf{0.220 ± 0.001} & \textbf{0.181 ± 0.001} & \textbf{0.534 ± 0.006} & \textbf{0.416 ± 0.007} & \textbf{0.449 ± 0.002} \\

\hline
Chronos-Bolt-Base & Zero-Shot 
& 0.339 & 0.184 & 0.168 & \textbf{0.371} & 0.177 & 0.145 & 0.772 & 0.698 & 0.696 \\

Chronos-Bolt-Base& FFT
& \textbf{0.329 ± 0.000} & \textbf{0.152 ± 0.001} & \textbf{0.138 ± 0.003} & 0.403 ± 0.009 & \textbf{0.171 ± 0.001} & \textbf{0.138 ± 0.001} & \textbf{0.580 ± 0.014} & \textbf{0.452 ± 0.010} & \textbf{0.402 ± 0.010} \\

\hline
TimesFM-2.0 & Zero-Shot 
& \textbf{0.326} & 0.177 & 0.154 & 0.467 & \textbf{0.251} & 0.224 & 0.579 & \textbf{0.627} & \textbf{0.667} \\

TimesFM-2.0 & FFT
& \textbf{0.326 ± 0.000} & \textbf{0.170 ± 0.001} & \textbf{0.142 ± 0.001} & \textbf{0.433 ± 0.002} & \textbf{0.251 ± 0.002} & \textbf{0.223 ± 0.001} & \textbf{0.559 ± 0.004} & 0.639 ± 0.017 & 0.698 ± 0.009
\\

 \hline
&& \multicolumn{9}{c}{\textbf{$\mathrm{MAPE}_n$}} \\
\hline
Moirai-1.1-R-L & Zero-shot 
&\textbf{0.576} & 0.559 & 0.458 & 0.733 & \textbf{0.752} & \textbf{0.970} & 0.679 & 1.103 & 1.097 \\

Moirai-1.1-R-L & FFT 
& 0.583 ± 0.007 & \textbf{0.524 ± 0.007} & \textbf{0.449 ± 0.011} & \textbf{0.693 ± 0.003} & 0.833 ± 0.003 & 0.975 ± 0.008 & \textbf{0.572 ± 0.012} & \textbf{0.717 ± 0.020} & \textbf{0.595 ± 0.007} \\

\hline
Chronos-Bolt-Base & Zero-Shot 
& 0.615 & 0.652 & 0.626 & \textbf{0.614} & \textbf{0.632} & \textbf{0.785} & 0.958 & 1.787 & 1.882 \\

Chronos-Bolt-Base& FFT
& \textbf{0.530 ± 0.003} & \textbf{0.560 ± 0.016} & \textbf{0.403 ± 0.045} & 0.674 ± 0.030 & 0.677 ± 0.022 & 0.824 ± 0.007 & \textbf{0.662 ± 0.037} & \textbf{0.503 ± 0.019} & \textbf{0.553 ± 0.066} \\

\hline
TimesFM-2.0 & Zero-Shot 
& 0.611 & \textbf{0.506} & 0.531 & \textbf{0.767} & \textbf{0.488} & \textbf{0.709} & \textbf{0.673} & 1.399 & 1.474 \\

TimesFM-2.0 & FFT
& \textbf{0.601 ± 0.005} & 0.509 ± 0.001 & \textbf{0.471 ± 0.007} & 0.850 ± 0.026 & 0.532 ± 0.009 & 0.748 ± 0.010 & 0.716 ± 0.012 & \textbf{1.352 ± 0.075} & \textbf{1.354 ± 0.081} \\

\hline

\hline
\end{tabular}
}
\caption{Moyennes de $\mathrm{MWQL}_n$ et $\mathrm{MAPE}_n$ selon les domaines (Transport, Nature et Web/CloudOps) et horizon (court/moyen/long) pour les trois modèles, en zéro-shot et en full fine-tuning (FFT).}
\label{tab:part2}
\end{table}

\begin{table}[t]
\centering
\small
\resizebox{\textwidth}{!}{
\begin{tabular}{ll|c|ccc|c|c}
\hline
\textbf{Modèle} & \textbf{Config} &
\textbf{Econ/Fin} & \multicolumn{3}{c|}{\textbf{Énergie}} & \textbf{Santé} & \textbf{Ventes} \\
\cline{3-8}
&& Short & Short & Medium & Long & Short & Short \\
\hline
&& \multicolumn{6}{c}{\textbf{$\mathrm{MWQL}_n$}} \\
\hline
Moirai-1.1-R-L & Zero-shot 
& 0.943 & 0.586 & 0.333 & 0.300 & 0.544 & 0.525 \\

Moirai-1.1-R-L & FFT 
& \textbf{0.715 ± 0.002} & \textbf{0.564 ± 0.007} & \textbf{0.280 ± 0.001} & \textbf{0.245 ± 0.001} & \textbf{0.462 ± 0.002} & \textbf{0.387 ± 0.001} \\

\hline
Chronos-Bolt-Base & Zero-Shot 
& 0.601 &       \textbf{0.525} & \textbf{0.286} & \textbf{0.259} & 0.483 &      0.384 \\

Chronos-Bolt-Base & FFT 
&\textbf{0.598 ± 0.002} & 0.588 ± 0.018 & 0.309 ± 0.006 & 0.289 ± 0.001 & \textbf{0.459 ± 0.012} & \textbf{0.382 ± 0.001} \\
\hline

TimesFM-2.0 & Zero-Shot 
& 0.660 & 0.596 & 0.301 & 0.277 & 0.407 & 0.377 \\
TimesFM-2.0 & FFT 
& \textbf{0.620 ± 0.004} & \textbf{0.573 ± 0.002} & \textbf{0.300 ± 0.012} & \textbf{0.274 ± 0.002} & \textbf{0.390 ± 0.001} & \textbf{0.374 ± 0.001} \\

\hline
&& \multicolumn{6}{c}{\textbf{$\mathrm{MAPE}_n$}} \\
\hline
Moirai-1.1-R-L & Zero-shot 
& 1.128 &             \textbf{ 0.938} & 0.991& \textbf{1.249} & 0.716 & 0.766 \\

Moirai-1.1-R-L & FFT 
& \textbf{0.855 ± 0.002} & 1.003 ± 0.016 & \textbf{0.957 ± 0.009} & 1.990 ± 0.054 & \textbf{0.585 ± 0.002} & \textbf{0.737 ± 0.001} \\

\hline
Chronos-Bolt-Base & Zero-Shot 
& \textbf{0.663} & \textbf{1.065} & \textbf{0.837} & \textbf{0.932} & 0.650 & \textbf{0.681} \\

Chronos-Bolt-Base & FFT
& 0.674 ± 0.001 & 1.141 ± 0.028 & 1.044 ± 0.042 & 1.517 ± 0.045 & \textbf{0.577 ± 0.008} & 0.736 ± 0.007 \\
\hline
TimesFM-2.0 & Zero-Shot 
& 0.725 & \textbf{1.131} & \textbf{1.225} & \textbf{1.575} & 0.577 & \textbf{0.635} \\

TimesFM-2.0 & FFT
& \textbf{0.708 ± 0.005} & 1.136 ± 0.007 & 1.315 ± 0.089 &1.920 ± 0.086 & \textbf{0.556 ± 0.006} & 0.706 ± 0.002 \\

\hline

\end{tabular}
}
\caption{Moyennes de $\mathrm{MWQL}_n$ et $\mathrm{MAPE}_n$) selon les domaines ({Économie/Finance}, {Énergie}, {Santé} et {Ventes}) et horizon (court/moyen/long) pour les trois modèles, en zéro-shot et en full fine-tuning (FFT).}
\label{tab:part1}
\end{table}

\section{Conclusion et perspective}\label{sec:conc}

Dans cet article, nous avons présenté une revue des modèles de fondation pour la prévision de séries temporelles, en les analysant sous plusieurs angles : architecture, pré-entraînement et adaptation par ajustement. Nous avons ensuite étudié l’effet de l’ajustement de ces modèles sur des jeux de données cibles, montrant que, avec un choix approprié des hyperparamètres, cette étape permet d’améliorer significativement les performances zero-shot, notamment pour les prévisions à long terme, où les modèles zero-shot tendent à être moins performants.

Le domaine des modèles de fondation pour la prévision de séries temporelles est très récent et présente encore de nombreux défis. D’une part, contrairement au NLP, les séries temporelles restent peu accessibles et souvent conservées par les grandes entreprises industrielles, ce qui limite la disponibilité de jeux de données vastes et diversifiés pour la recherche. D’autre part, les modèles de fondation appliqués aux séries temporelles manquent encore d’études théoriques et empiriques permettant de mieux comprendre leur comportement en matière de généralisation et de compression de l’information, d’autant plus que les séries temporelles présentent une forte hétérogénéité, rendant l’apprentissage d’un préviseur universel particulièrement difficile.

\bibliographystyle{rnti}
\bibliography{biblioForArxiv}
\appendix

\section{Informations complémentaires sur les jeux de données utilisés}

\subsection{Caractéristiques des séries temporelles selon les différents domaines}

Dans cette section, nous présentons une analyse détaillée des caractéristiques statistiques des séries temporelles issues des jeux de données train/test du benchmark GIFTEval~\cite{aksu2024gifteval}. Cette analyse vise à mieux comprendre les particularités structurelles des données selon les différents domaines étudiés (finance, santé, énergie, transport, etc.).  
Pour cela, nous reprenons trois mesures fondamentales permettant d’étudier les caractéristiques et la prévisibilité d’une série temporelle à partir de son historique : la \textbf{tendance}, la \textbf{saisonnalité} et l’\textbf{entropie}.

\paragraph{Tendance}  
La tendance mesure l’évolution globale d’une série temporelle au fil du temps (hausse, baisse ou stabilité). Elle est obtenue via une décomposition de la série en composantes de tendance, de saisonnalité et de résidu, puis en comparant la variance du composant résiduel à la variance combinée des composants de tendance et de résidu. La force de la tendance est définie comme suit :
\[
\text{trend} = 1 - \frac{\operatorname{Var}(e_t)}{\operatorname{Var}(f_t + e_t)},
\]
où \(f_t\) est la composante de tendance et \(e_t\) le résidu.  
Des valeurs proches de 1 indiquent une tendance marquée, comme c’est souvent le cas pour les séries économiques ou financières. À l’inverse, certaines séries provenant de domaines comme le transport présentent une tendance plus faible.

\paragraph{Saisonnalité}  
La force de la saisonnalité traduit l’existence de motifs répétitifs à intervalles réguliers—par exemple des cycles journaliers (Ventes, énergie) ou annuels (finance).  
Elle est calculée pour chaque composante saisonnière \(s_{i,t}\) selon :
\[
\text{seasonal\_strength}_i = 1 - \frac{\operatorname{Var}(e_t)}{\operatorname{Var}(s_{i,t} + e_t)}.
\]
Une valeur élevée indique une présence forte de motifs récurrents, rendant la série plus structurée.

\paragraph{Entropie / Prévisibilité}  
L’entropie spectrale mesure la complexité ou le niveau de bruit d’une série temporelle :
\[
\text{Entropy} = -\int_{-\pi}^{\pi} \hat{f}(\lambda) \log \hat{f}(\lambda) \, d\lambda,
\]
où \(\hat{f}(\lambda)\) est l’estimation de la densité spectrale.  
Une faible entropie traduit une structure forte et une série facilement prédictible, tandis qu’une entropie élevée caractérise un comportement irrégulier ou bruité.  

\begin{figure}[t]
    \centering
    \includegraphics[width=10cm]{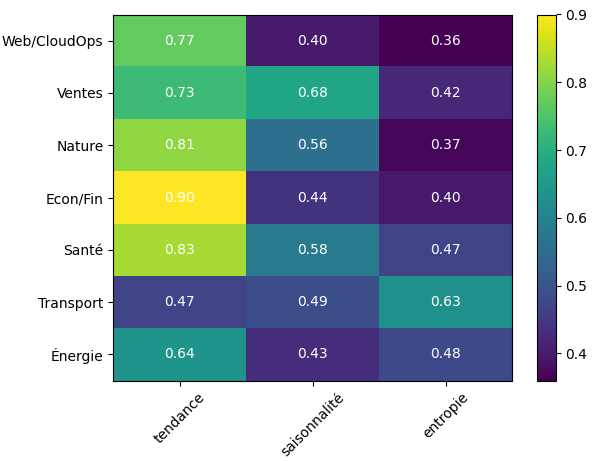}
    \caption{Cartes thermiques représentant les valeurs moyennes de trois caractéristiques de séries temporelles selon les différents domaines du jeu de données.}
    \label{fig:ts_heatmaps}
\end{figure}

\paragraph{}  
La Figure~\ref{fig:ts_heatmaps} illustre les valeurs moyennes de ces trois caractéristiques pour chacun des domaines considérés, mettant en évidence des disparités structurelles importantes. Ces observations confirment l’importance d’évaluer les modèles de prévision sur un ensemble diversifié de conditions temporelles, comme le propose GIFTEval sur la figure \ref{fig:ts_heatmaps}.

\subsection{Jeux de données utilisés et tailles associées}

Le tableau ~\ref{tab:dataset} présentons l’ensemble des jeux de données employés dans notre étude, ainsi que leurs caractéristiques principales, notamment leur taille, leur domaine, leur fréquence et la longueur moyenne des séries temporelles. Ces informations permettent de mieux comprendre la diversité et la complexité des données sur lesquelles les modèles sont évalués.

\begin{table}
\resizebox{\textwidth}{!}{%
\begin{tabular}{cccccccccccccccc}
\hline
\textbf{} &
  \textbf{} &
  \textbf{} &
  \textbf{} &
  \textbf{} &
  \multicolumn{3}{c}{\textbf{Longueur des séries}} &
  \textbf{} &
  \textbf{} &
  \multicolumn{2}{c}{\textbf{Court terme}} &
  \multicolumn{2}{c}{\textbf{Moyen terme}} &
  \multicolumn{2}{c}{\textbf{Long terme}} \\ \hline
\textbf{Jeu de données} &
  \textbf{Source} &
  \textbf{Domaine[Taille]} &
  \textbf{Fréquence} &
  \textbf{\# Séries} &
  \textbf{Moy.} &
  \textbf{Min} &
  \textbf{Max} &
  \textbf{\# Obs.} &
  \textbf{Variables cibles} &
  \textbf{Horizon (C)} &
  \textbf{Fenêtres de Test} &
  \textbf{Horizon (M)} &
  \textbf{Fenêtres de Test} &
  \textbf{Horizon (L)} &
  \textbf{Fenêtres de Test} \\
Jena Weather &
  Autoformer~\citep{Wu2021AutoformerDT} &
  Nature[Grand] &
  10T &
  1 &
  52,704 &
  52,704 &
  52,704 &
  52,704 &
  21 &
  48 &
  20 &
  480 &
  11 &
  720 &
  8 \\
Jena Weather &
  Autoformer~\citep{Wu2021AutoformerDT} &
  Nature[Moyen] &
  H &
  1 &
  8,784 &
  8,784 &
  8,784 &
  8,784 &
  21 &
  48 &
  19 &
  480 &
  2 &
  720 &
  2 \\
Jena Weather &
  Autoformer~\citep{Wu2021AutoformerDT} &
  Nature[Petit] &
  D &
  1 &
  366 &
  366 &
  366 &
  366 &
  21 &
  30 &
  2 &
   &
   &
   &
   \\
BizITObs - Application &
  AutoMixer~\citep{10.1609/aaai.v38i21.30336} &
  Web/CloudOps[Petit] &
  10S &
  1 &
  8,834 &
  8,834 &
  8,834 &
  8,834 &
  2 &
  60 &
  15 &
  600 &
  2 &
  900 &
  1 \\
BizITObs - Service &
  AutoMixer~\citep{10.1609/aaai.v38i21.30336} &
  Web/CloudOps[Moyen] &
  10S &
  21 &
  8,835 &
  8,835 &
  8,835 &
  185,535 &
  2 &
  60 &
  15 &
  600 &
  2 &
  900 &
  1 \\
BizITObs - L2C &
  AutoMixer~\citep{10.1609/aaai.v38i21.30336} &
  Web/CloudOps[Moyen] &
  5T &
  1 &
  31,968 &
  31,968 &
  31,968 &
  31,968 &
  7 &
  48 &
  20 &
  480 &
  7 &
  720 &
  5 \\
BizITObs - L2C &
  AutoMixer~\citep{10.1609/aaai.v38i21.30336} &
  Web/CloudOps[Petit] &
  H &
  1 &
  2,664 &
  2,664 &
  2,664 &
  2,664 &
  7 &
  48 &
  6 &
  480 &
  1 &
  720 &
  1 \\
Bitbrains - Fast Storage &
  Grid Workloads Archive~\citep{10.1109/CCGrid.2015.60} &
  Web/CloudOps[Grand] &
  5T &
  1,250 &
  8,640 &
  8,640 &
  8,640 &
  10,800,000 &
  2 &
  48 &
  18 &
  480 &
  2 &
  720 &
  2 \\
Bitbrains - Fast Storage &
  Grid Workloads Archive~\citep{10.1109/CCGrid.2015.60} &
  Web/CloudOps[Grand] &
  H &
  1,250 &
  721 &
  721 &
  721 &
  901,250 &
  2 &
  48 &
  2 &
   &
   &
   &
   \\
Bitbrains - rnd &
  Grid Workloads Archive~\citep{10.1109/CCGrid.2015.60} &
  Web/CloudOps[Grand] &
  5T &
  500 &
  8,640 &
  8,640 &
  8,640 &
  4,320,000 &
  2 &
  48 &
  18 &
  480 &
  2 &
  720 &
  2 \\
Bitbrains - rnd &
  Grid Workloads Archive~\citep{10.1109/CCGrid.2015.60} &
  Web/CloudOps[Grand] &
  H &
  500 &
  720 &
  720 &
  720 &
  360,000 &
  2 &
  48 &
  2 &
   &
   &
   &
   \\

ETT1 &
  Informer~\citep{Zhou2020InformerBE} &
  Energy[Moyen] &
  15T &
  1 &
  69,680 &
  69,680 &
  69,680 &
  69,680 &
  7 &
  48 &
  20 &
  480 &
  15 &
  720 &
  10 \\
ETT1 &
  Informer~\citep{Zhou2020InformerBE} &
  Energy[Moyen] &
  H &
  1 &
  17,420 &
  17,420 &
  17,420 &
  17,420 &
  7 &
  48 &
  20 &
  480 &
  4 &
  720 &
  3 \\
ETT1 &
  Informer~\citep{Zhou2020InformerBE} &
  Energy[Petit] &
  D &
  1 &
  725 &
  725 &
  725 &
  725 &
  7 &
  30 &
  3 &
   &
   &
   &
   \\
ETT1 &
  Informer~\citep{Zhou2020InformerBE} &
  Energy[Petit] &
  W-THU &
  1 &
  103 &
  103 &
  103 &
  103 &
  7 &
  8 &
  2 &
   &
   &
   &
   \\
ETT2 &
  Informer~\citep{Zhou2020InformerBE} &
  Energy[Moyen] &
  15T &
  1 &
  69,680 &
  69,680 &
  69,680 &
  69,680 &
  7 &
  48 &
  20 &
  480 &
  15 &
  720 &
  10 \\
ETT2 &
  Informer~\citep{Zhou2020InformerBE} &
  Energy[Moyen] &
  H &
  1 &
  17,420 &
  17,420 &
  17,420 &
  17,420 &
  7 &
  48 &
  20 &
  480 &
  4 &
  720 &
  3 \\
ETT2 &
  Informer~\citep{Zhou2020InformerBE} &
  Energy[Petit] &
  D &
  1 &
  725 &
  725 &
  725 &
  725 &
  7 &
  30 &
  3 &
   &
   &
   &
   \\
ETT2 &
  Informer~\citep{Zhou2020InformerBE} &
  Energy[Petit] &
  W-THU &
  1 &
  103 &
  103 &
  103 &
  103 &
  7 &
  8 &
  2 &
   &
   &
   &
   \\
Loop Seattle &
  LibCity~\citep{Wang2023TowardsEA} &
  Transport[Grand] &
  5T &
  323 &
  105,120 &
  105,120 &
  105,120 &
  33,953,760 &
  1 &
  48 &
  20 &
  480 &
  20 &
  720 &
  15 \\
Loop Seattle &
  LibCity~\citep{Wang2023TowardsEA} &
  Transport[Grand] &
  H &
  323 &
  8,760 &
  8,760 &
  8,760 &
  2,829,480 &
  1 &
  48 &
  19 &
  480 &
  2 &
  720 &
  2 \\
Loop Seattle &
  LibCity~\citep{Wang2023TowardsEA} &
  Transport[Petit] &
  D &
  323 &
  365 &
  365 &
  365 &
  117,895 &
  1 &
  30 &
  2 &
   &
   &
   &
   \\
SZ-Taxi &
  LibCity~\citep{Wang2023TowardsEA} &
  Transport[Moyen] &
  15T &
  156 &
  2,976 &
  2,976 &
  2,976 &
  464,256 &
  1 &
  48 &
  7 &
  480 &
  1 &
  720 &
  1 \\
SZ-Taxi &
  LibCity~\citep{Wang2023TowardsEA} &
  Transport[Moyen] &
  H &
  156 &
  744 &
  744 &
  744 &
  116,064 &
  1 &
  48 &
  2 &
   &
   &
   &
   \\
M\_DENSE &
  LibCity~\citep{Wang2023TowardsEA} &
  Transport[Grand] &
  H &
  30 &
  17,520 &
  17,520 &
  17,520 &
  525,600 &
  1 &
  48 &
  20 &
  480 &
  4 &
  720 &
  3 \\
M\_DENSE &
  LibCity~\citep{Wang2023TowardsEA} &
  Transport[Petit] &
  D &
  30 &
  730 &
  730 &
  730 &
  21,900 &
  1 &
  30 &
  3 &
   &
   &
   &
   \\
Solar &
  LSTNet~\citep{Lai2017ModelingLA} &
  Energy[Grand] &
  10T &
  137 &
  52,560 &
  52,560 &
  52,560 &
  7,200,720 &
  1 &
  48 &
  20 &
  480 &
  11 &
  720 &
  8 \\
Solar &
  LSTNet~\citep{Lai2017ModelingLA} &
  Energy[Grand] &
  H &
  137 &
  8,760 &
  8,760 &
  8,760 &
  1,200,120 &
  1 &
  48 &
  19 &
  480 &
  2 &
  720 &
  2 \\
Solar &
  LSTNet~\citep{Lai2017ModelingLA} &
  Energy[Petit] &
  D &
  137 &
  365 &
  365 &
  365 &
  50,005 &
  1 &
  30 &
  2 &
   &
   &
   &
   \\
Solar &
  LSTNet~\citep{Lai2017ModelingLA} &
  Energy[Petit] &
  W-FRI &
  137 &
  52 &
  52 &
  52 &
  7,124 &
  1 &
  8 &
  1 &
   &
   &
   &
   \\
Hierarchical Sales &
  \citet{Mancuso2020AML} &
  Sales[Moyen] &
  D &
  118 &
  1,825 &
  1,825 &
  1,825 &
  215,350 &
  1 &
  30 &
  7 &
   &
   &
   &
   \\
Hierarchical Sales &
  \citet{Mancuso2020AML} &
  Sales[Petit] &
  W-WED &
  118 &
  260 &
  260 &
  260 &
  30,680 &
  1 &
  8 &
  4 &
   &
   &
   &
   \\

M4 Monthly &
  Monash~\citep{godahewa2021monash} &
  Econ/Fin[Grand] &
  M &
  48,000 &
  234 &
  60 &
  2,812 &
  11,246,411 &
  1 &
  18 &
  1 &
   &
   &
   &
   \\
M4 Weekly &
  Monash~\citep{godahewa2021monash} &
  Econ/Fin[Moyen] &
  W-SUN &
  359 &
  1,035 &
  93 &
  2,610 &
  371,579 &
  1 &
  13 &
  1 &
   &
   &
   &
   \\
M4 Daily &
  Monash~\citep{godahewa2021monash} &
  Econ/Fin[Grand] &
  D &
  4,227 &
  2,371 &
  107 &
  9,933 &
  10,023,836 &
  1 &
  14 &
  1 &
   &
   &
   &
   \\
M4 Hourly &
  Monash~\citep{godahewa2021monash} &
  Econ/Fin[Moyen] &
  H &
  414 &
  902 &
  748 &
  1,008 &
  373,372 &
  1 &
  48 &
  2 &
   &
   &
   &
   \\
Hospital &
  Monash~\citep{godahewa2021monash} &
  Healthcare[Moyen] &
  M &
  767 &
  84 &
  84 &
  84 &
  64,428 &
  1 &
  12 &
  1 &
   &
   &
   &
   \\
COVID Deaths &
  Monash~\citep{godahewa2021monash} &
  Healthcare[Moyen] &
  D &
  266 &
  212 &
  212 &
  212 &
  56,392 &
  1 &
  30 &
  1 &
   &
   &
   &
   \\
US Births &
  Monash~\citep{godahewa2021monash} &
  Healthcare[Petit] &
  D &
  1 &
  7,305 &
  7,305 &
  7,305 &
  7,305 &
  1 &
  30 &
  20 &
   &
   &
   &
   \\
US Births &
  Monash~\citep{godahewa2021monash} &
  Healthcare[Petit]  &
  W-TUE &
  1 &
  1,043 &
  1,043 &
  1,043 &
  1,043 &
  1 &
  8 &
  14 &
   &
   &
   &
   \\
US Births &
  Monash~\citep{godahewa2021monash} &
  Healthcare[Petit]  &
  M &
  1 &
  240 &
  240 &
  240 &
  240 &
  1 &
  12 &
  2 &
   &
   &
   &
   \\
Saugeen &
  Monash~\citep{godahewa2021monash} &
  Nature[Petit]  &
  D &
  1 &
  23,741 &
  23,741 &
  23,741 &
  23,741 &
  1 &
  30 &
  20 &
   &
   &
   &
   \\
Saugeen &
  Monash~\citep{godahewa2021monash} &
  Nature[Petit]  &
  W-THU &
  1 &
  3,391 &
  3,391 &
  3,391 &
  3,391 &
  1 &
  8 &
  20 &
   &
   &
   &
   \\
Saugeen &
  Monash~\citep{godahewa2021monash} &
  Nature[Petit]  &
  M &
  1 &
  780 &
  780 &
  780 &
   &
  1 &
  12 &
  7 &
   &
   &
   &
   \\
Temperature Rain &
  Monash~\citep{godahewa2021monash} &
  Nature[Grand]  &
  D &
  32,072 &
  725 &
  725 &
  725 &
  780 &
  1 &
  30 &
  3 &
   &
   &
   &
   \\
KDD Cup 2018 &
  Monash~\citep{godahewa2021monash} &
  Nature[Grand]  &
  H &
  270 &
  10,898 &
  9,504 &
  10,920 &
  2,942,364 &
  1 &
  48 &
  20 &
  480 &
  2 &
  720 &
  2 \\
KDD Cup 2018 &
  Monash~\citep{godahewa2021monash} &
  Nature[Moyen]  &
  D &
  270 &
  455 &
  396 &
  455 &
  122,791 &
  1 &
  30 &
  2 &
   &
   &
   &
   \\
Car Parts &
  Monash~\citep{godahewa2021monash} &
  Sales[Moyen]  &
  M &
  2,674 &
  51 &
  51 &
  51 &
  136,374 &
  1 &
  12 &
  1 &
   &
   &
   &
   \\
Electricity &
  UCI ML Archive~\citep{electricityloaddiagrams20112014_321} &
  Energy[Grand] &
  15T &
  370 &
  140,256 &
  140,256 &
  140,256 &
  51,894,720 &
  1 &
  48 &
  20 &
  480 &
  20 &
  720 &
  20 \\
Electricity &
  UCI ML Archive~\citep{electricityloaddiagrams20112014_321} &
  Energy[Grand] &
  H &
  370 &
  35,064 &
  35,064 &
  35,064 &
  12,973,680 &
  1 &
  48 &
  20 &
  480 &
  8 &
  720 &
  5 \\
Electricity &
  UCI ML Archive~\citep{electricityloaddiagrams20112014_321} &
  Energy[Grand] &
  D &
  370 &
  1,461 &
  1,461 &
  1,461 &
  540,570 &
  1 &
  30 &
  5 &
   &
   &
   &
   \\
Electricity &
  UCI ML Archive~\citep{electricityloaddiagrams20112014_321} &
  Energy[Moyen] &
  W-FRI &
  370 &
  208 &
  208 &
  208 &
  76,960 &
  1 &
  8 &
  3 &
   &
   &
   &
   \\ \hline
\end{tabular}%
}
\caption{Statistiques individuelles de chaque jeu de données du benchmark, incluant notamment la taille de chacun d’eux.}
\label{tab:dataset}

\end{table}

\end{document}